\documentclass[lettersize,journal，11pt]{IEEEtran}
\usepackage{amsmath,amsfonts}
\usepackage{algorithm}
\usepackage{algpseudocode}
\usepackage{array}
\usepackage[caption=false,font=normalsize,labelfont=sf,textfont=sf]{subfig}
\usepackage{textcomp}
\usepackage{stfloats}
\usepackage{url}
\usepackage{verbatim}
\usepackage{graphicx}
\usepackage{xcolor}
\usepackage{caption}
\usepackage[utf8]{inputenc}
\usepackage[english]{babel}

\usepackage[square,sort,comma,numbers]{natbib}

\usepackage{booktabs}
\usepackage{multirow} 
\usepackage{pifont}

\renewcommand{\arraystretch}{1.25}
\begin{document}

\title{Dynamic Disentangled Fusion Network for RGBT Tracking}

\author{ Chenglong Li, Tao Wang, Zhaodong Ding, Yun Xiao, Jin Tang
\thanks{This work is supported in part by the National Natural Science Foundation of China under Grant 62376004, in part by the Natural Science Foundation of Anhui Province under Grant 2208085J18.}
\thanks{
Chenglong Li, Tao Wang, Zhaodong Ding, and Yun Xiao are with Information Materials and Intelligent Sensing Laboratory of Anhui Province, Anhui Provincial Key Laboratory of Multimodal Cognitive Computation, School of Artificial Intelligence, Anhui University, Hefei 230601, China. 
(e-mail: lcl1314@foxmail.com; WA23301097@stu.ahu.edu.cn; zhaodongding\_ah@163.com; xiaoyun@ahu.edu.cn)}
\thanks{Jin Tang is with Anhui Provincial Key Laboratory of Multimodal Cognitive Computation, School of Computer Science and Technology, Anhui University, Hefei 230601, China. ( tangjin@ahu.edu.cn)
}}

\markboth{Journal of \LaTeX\ Class Files,~Vol.~14, No.~8, August~2021}%
{Shell \MakeLowercase{\textit{et al.}}: A Sample Article Using IEEEtran.cls for IEEE Journals}

\maketitle

\begin{abstract}
RGBT tracking usually suffers from various challenging factors of low resolution, similar appearance, extreme illumination, thermal crossover and occlusion, to name a few. 
Existing works often study complex fusion models to handle challenging scenarios, but can not well adapt to various challenges, which might limit tracking performance. 
To handle this problem, we propose a novel Dynamic Disentangled Fusion Network called DDFNet, which disentangles the fusion process into several dynamic fusion models via the challenge attributes to adapt to various challenging scenarios, for robust RGBT tracking. 
In particular, we design six attribute-based fusion models to integrate RGB and thermal features under the six challenging scenarios respectively.
Since each fusion model is to deal with the corresponding challenges, such disentangled fusion scheme could increase the fusion capacity without the dependence on large-scale training data.
Considering that every challenging scenario also has different levels of difficulty, we propose to optimize the combination of multiple fusion units to form each attribute-based fusion model in a dynamic manner, which could well adapt to the difficulty of the corresponding challenging scenario.
To address the issue that which fusion models should be activated in the tracking process, we design an adaptive aggregation fusion module to integrate all features from attribute-based fusion models in an adaptive manner with a three-stage training algorithm.
In addition, we design an enhancement fusion module to further strengthen the aggregated feature and modality-specific features. 
Experimental results on benchmark datasets demonstrate the effectiveness of our DDFNet against other state-of-the-art methods.
\end{abstract}


\section{Introduction}
\IEEEPARstart{R}{GBT} RGBT tracking is to track the specific target in a RGB (visible light) and TIR (thermal infrared) video sequence pair according to its state including position and size given in the initial frame. 
By employing the complementary advantages of RGB and TIR modalities, it achieves robust tracking in various harsh conditions such as dark environments and bad weather, and has been widely used in many practical applications like surveillance security, intelligent UAV system and autonomous driving. However, RGBT tracking is still a difficult task as it usually suffers from various challenging factors of low resolution, similar appearance,
extreme illumination, thermal crossover and occlusion, to name a few.

Existing RGBT tracking works~\cite{QAT2023, Fan2024, Li2023sensors,lu2024after, Zhu2024} often try to study various fusion models to overcome the effects of various challenges. 
Some works~\cite{lu2024after,Fan2024,TATrack} focus on the effective fusion of features from two modalities for robust tracking performance.
For instance, Fan et al.~\cite{Fan2024} propose a joint-modality query fusion network to achieve the mutually reinforcing of feature extraction and fusion, in which intra-modal feature extraction and inter-modal fusion are coupled together and mutually promoted by joint-modality queries. 
Although these works can alleviate the interference of low-quality modality information, single fusion models struggle to address the negative impact of various challenging scenarios on multimodal fusion.
Unlike the aforementioned methods, some works~\cite{li2020rgbt,liu2024rgbt, Zhang2021rgbt} attempt to utilize the disentangled representation learning to learn target features under different challenging scenarios, and then achieve robust feature representations through adaptive fusion.
%
For instance, Liu et al.~\cite{liu2024rgbt} disentangle the target appearance through five different challenge-specific branches, and then aggregate all features to interact and form robust appearance representations. 
However, different challenging scenarios have varying levels of difficulty and are dynamically changing. These methods often use fixed structural branches to tackle various challenges, making it difficult to effectively adapt to real-world challenging scenarios. In addition, these methods focus on the target appearance modeling under certain challenges, but neglect the effective modeling of multimodal fusion, which is the key issue in RGBT tracking. 

%
%

\begin{figure}
\centering
\includegraphics[width=0.48\textwidth]{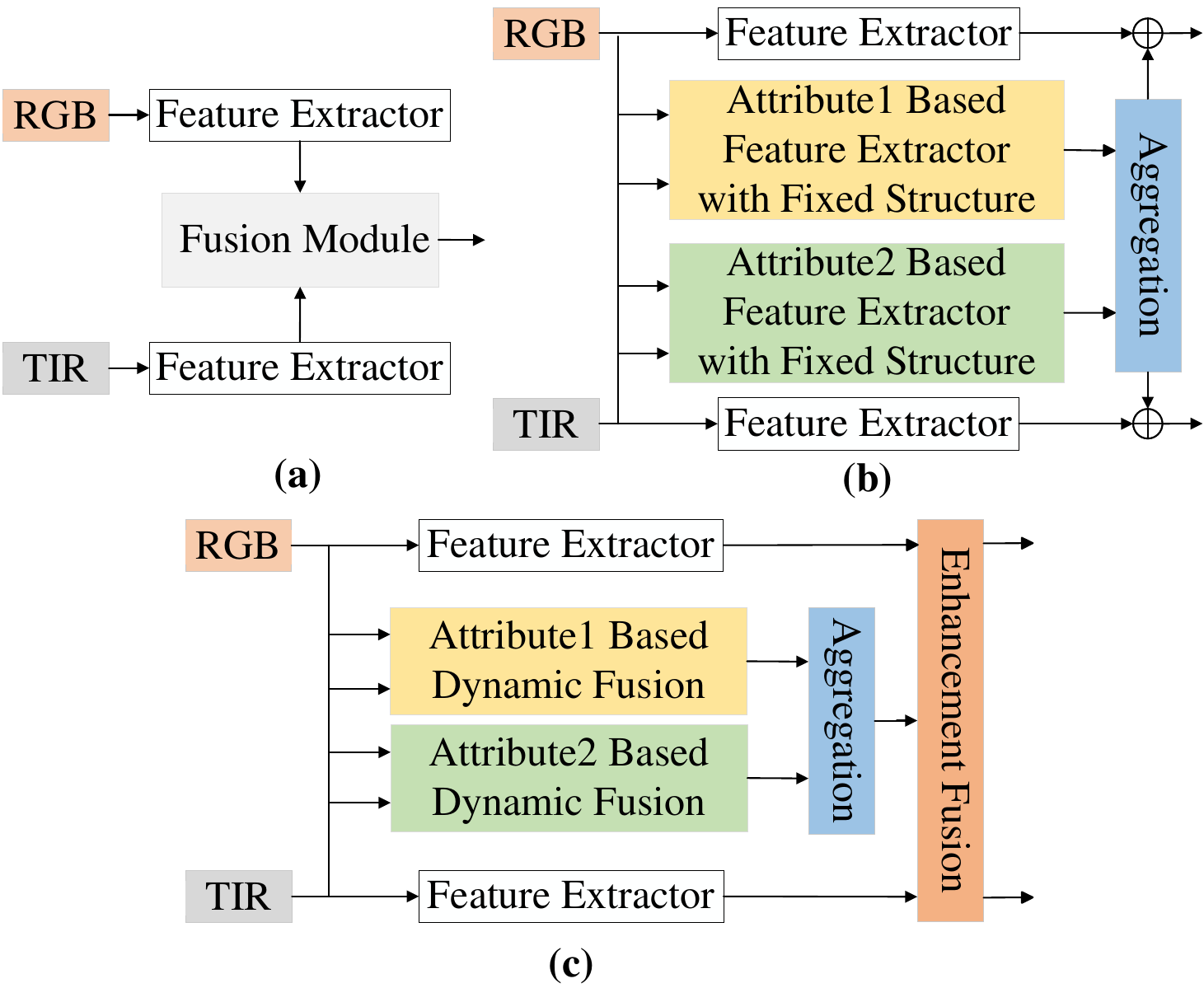}
\caption{ Comparison of our dynamic disentangled fusion model with existing methods. The common fusion models (a) tend to design a complex single-branch fusion network. In the existing attribute-based appearance disentanglement models (b) which extract appearance features under certain attributes and then
perform feature fusion, but each branch has a fixed structure. In our DDFNet (c), each dynamic fusion branch dynamically selects fusion units to compose the fusion structure according to the challenge scenario, and this design can better perform effective fusion under the corresponding challenge attributes.
}
\label{introduction}
\end{figure}

To solve these problems, we propose a novel Dynamic Disentangled Fusion Network (DDFNet),  which disentangles the fusion process via six attribute-based dynamic fusion models to adapt to various challenging scenarios, for robust RGBT tracking. 
As shown in Figure~\ref{introduction}, it shows the major differences of our fusion model over existing approaches~\cite{li2020rgbt,liu2024rgbt,Zhang2021rgbt}.

On the one hand, to mitigate the adverse effects of different challenging scenarios on multimodal fusion, we disentangle the fusion process via six challenge attributes including occlusion (OCC), low resolution (LR), similar appearance (SV), extreme illumination (EI), thermal crossover (TC) and general attribute (GEN)~\cite{Zhang2021rgbt}, whereas the introduction of general attribute to better cope with the challenges beyond the above five challenge attributes. 
Unlike disentangling the target representations for appearance modeling in existing methods~\cite{li2020rgbt,liu2024rgbt}, we disentangle the fusion process via different challenge attributes to achieve effective modeling of multimodal fusion for the adaptation to various challenging scenarios.
It is worth noting that each fusion model is only responsible for a specific challenge attribute, we can employ a small number of parameters for fusion model design.
Additionally, each fusion model can be trained independently using small-scale training data related to their respective attributes without the dependence on large-scale training data.

Since it is unknown which fusion models should be activated during the tracking phase, we design an adaptive aggregation fusion module based on  SKNet~\cite{Li2019SelectiveKN} to adaptively aggregate the features from different attribute-specific fusion models.
During the training stage, we first learn the attribute-specific fusion models by employing the corresponding training subsets, and then train the adaptive aggregation fusion module using entire dataset. Such scheme allows this module to automatically perceive the appeared attributes of input data and thus to adaptively aggregate the features from the appeared attribute branches. 
To further enhance multimodal representations, different from Transformer-based enhancement fusion~\cite{Xiao2022rgbt}, we design a lightweight enhancement fusion module to perform the comprehensive information interaction between modalities by employing aggregated features to guide the enhancement of the features from both modalities. 
In specific, we first use the aggregated features to generate spatial weights through a convolution layer and an activation function, and then apply these spatial weights to enhance the features of both modalities. 
Afterward, the aggregated features are fused with the features of each modality separately to achieve comprehensive multimodal information interaction.

On the other hand, considering that each challenging scenario has varying levels of difficulty, we design the dynamic fusion structure based on some base fusion units to adapt to dynamic challenging environments. 
Current attribute-based tracking networks~\cite{li2020rgbt,liu2024rgbt, Zhang2021rgbt} often employ the fixed fusion architecture for each challenging attribute, and lack the adaptability to dynamic challenging scenarios. To handle this problem, we optimize the combination of multiple fusion units to form each attribute-based fusion model in a dynamic manner, which can well adapt to the difficulty of the corresponding challenging scenario. 
In particular, each dynamic fusion model consists of two spatial-channel fusion units and a selective fusion unit.
The spatial-channel fusion unit is used to enhance multimodal features and suppress target-irrelevant information by exploiting the spatial and channel attentions.
To achieve effective fusion of complementary information from both modalities, we introduce the selective fusion unit to predict the channel-level weights.
In addition, each unit of the dynamic fusion model embeds a simple router to predict the combination weights of fused features. 
In this way, our dynamic fusion model can build the suitable fusion structure according to the different levels of challenge difficulty.

We implement a dual-stream hierarchical architecture to progressively integrate the dynamic disentangled fusion modules. Comprehensive experiments are performed on four RGBT tracking datasets, including GTOT~\cite{li2016rgbt}, RGBT210~\cite{li2017rgbt}, RGBT234~\cite{li2019rgbt} and LasHeR~\cite{li2022rgbt}. The experimental results demonstrate that our method achieves promising performance against state-of-the-art methods. 
The contributions of this paper are summarized as follows.

\begin{itemize}
\item  We propose a novel Dynamic Disentangled Fusion Network to handle various challenging scenarios by disentangling the fusion process via six challenge attributes in RGBT tracking. Each attribute-based fusion model just needs to focus on the feature fusion under a certain challenging scenario, and can thus be efficiently trained using small-scale training data.

\item We design a dynamic fusion module for the effective fusion of each challenge attribute to adapt to the varying levels of challenge difficulty. It can optimize the combination of multiple fusion units to form each attribute-based fusion model in a dynamic manner, and different difficulty levels of challenge factors can thus be well handled.

\item We design an adaptive aggregation fusion module and a three-stage training algorithm to adaptively aggregate all attribute-based fusion features according to the challenging factors of input data. Although which fusion branches should be activated in the tracking process are unknown, we employ the three-stage training algorithm  to enable the aggregation fusion module to effectively suppress noisy features from unappeared attributes.

\item We design a lightweight enhancement fusion module that enables comprehensive information interaction between modalities by employing the aggregated features to guide the enhancement of the features from both modalities.

\end{itemize}

This work called DDFNet is an extension of our conference version APFNet~\cite{Xiao2022rgbt} with four main improvements. 
First, different from APFNet that uses the fixed structure for attribute-based fusion branches, our DDFNet introduces the dynamic fusion structure based on several fusion units and a router.
Second, considering the attributes such as scale variation and rapid motion require taking into account the changes in the target's appearance over time, we decide to exclude these two attributes in DDFNet, and add one new general challenge and two new specific challenges, i.e. low-resolution, and similar appearance.
Third, unlike employing Transformer-based enhancement fusion with high computational complexity in APFNet, our DDFNet introduces a lightweight CNN-based enhancement fusion module to employ the aggregated features to guide the enhancement of the features from both modalities.
Finally, to extract more robust target features, we replace the baseline network MDNet with the more powerful ToMP50 network in DDFNet, enabling us to fully utilize the information from both modalities.
Compared to APFNet, DDFNet achieves a significant boost in tracking performance, and the improvements in PR/SR are 1.1\%/2.6\%, 7.3\%/9.0\%, and 21.1\%/19.7\% on GTOT, RGBT234, and LasHeR datasets, respectively.

\section{Related Work}
In this section, we give a brief introduction to RGBT tracking methods and dynamic routing methods.
\subsection{RGBT Tracking Methods}
In the past few years, RGBT tracking has made remarkable progress as many algorithms with excellent performance are proposed. Some researchers focus on exploring the adequate fusion of multimodal information in RGBT tracking. For instance, 
Liu et al.~\cite{QAT2023} propose a quality-aware network that learns the reliability of different modalities based on a supervised approach and then uses the predicted reliability weights of different modalities to achieve adaptive fusion. 
Zhu et al.~\cite{Zhu2023VisualPM} fine-tune the base tracking model of RGB by using a prompt learning and inject thermal infrared modal information in the form of prompt messages to complete RGBT fusion.
Tang et al.~\cite{TANG2023101881} explore and compare the fusion of RGB and TIR information in multimodal target tracking, including pixel-level, feature-level, and decision-level fusion, and highlight the advantages of fusing multimodal information at the decision level. 
Cao et al.~\cite{BAT2024} propose a multimodal visual prompt tracking model based on a bi-directional adapter, where only a small number of parameters are fine-tuned. 
However, despite the impressive progress of these works, they are hard to model robust appearance representations of targets in some challenging scenarios.

To enhance target appearance representations in complex scenes even with small-scale training data, some researchers attempt to explore attribute-based representations to improve tracking robustness~\cite{qi2019,li2020rgbt, Zhang2021rgbt, liu2024rgbt}. 
Li et al.~\cite{li2020rgbt} first explore the potential of the attribute-based appearance disentanglement method in RGBT tracking, which contains parameter-independent branches and parameter-sharing branches to address the modality-specific and modality-shared challenges respectively.
Zhang et al.~\citep{Zhang2021rgbt} design different attribute branches to solve different attribute challenges comparing with~\cite{li2020rgbt}, especially introducing a common branch that can address a wide range of challenges, and propose an attribute-integrated network at channel-level and spatial-level to adaptively aggregate feature representations under different challenges.
%
%
Liu et al.~\citep{liu2024rgbt} try to design different branch structures based on the characteristics of different challenges so that the branches can better adapt to the specific challenges.
While these methods significantly improve tracking robustness in different challenging scenarios, the capability of fusion models to cope with the modeling of target representations under some key attributes is still inadequate.
Different from the above methods, we propose the attribute-based dynamic fusion branch that can build different fusion structures by combining different fusion units and adaptively selecting a fusion structure suitable for the challenging scenario of input data.

\subsection{Dynamic Routing Methods}
In recent years, dynamic networks have received increasing attention due to their powerful capabilities and have now been applied in various deep learning fields~\cite{Hu2023ADM, Li2023DynaMaskDM}.
In comparison to inference neural networks that search with a fixed structure, these networks can generate dynamic paths and select appropriate structures based on input information. 
For example, dynamic routing methods for network compression are achieved through techniques such as channel pruning~\cite{You2019Gate} or layer skipping~\cite{wang2018skipnet}.
In addition, semantic segmentation~\cite{Li2020LearningDR} and object detection~\cite{Song2020FineGrainedDH} utilize dynamic routing networks to fully capture the multi-scale features in input samples.
Wu et al.~\citep{Wu2017BlockDropDI} propose a method that dynamically selects which layers of a deep network to run during inference, optimizing computational cost while maintaining prediction accuracy.
Tsai et al.~\citep{Tsai2020MultimodalRI} dynamically adjust the weights between input modalities and output representations via multimodal routing for each input sample, thereby determining the relative importance of both individual modalities and cross-modality features.
Zeng et al.~\citep{ZENG2024107335} propose a feature-repair-based dynamic interaction network for multimodal sentiment analysis, aiming at addressing issues of signal noise and loss during the input phase, as well as low feature utilization efficiency during the modality fusion phase.

Although dynamic routing methods have made significant progress in these fields, further exploration is needed in RGBT tracking. 
Lu et al.~\cite{lu2024after} first use the dynamic routing method to optimize the fusion structure to adapt to the dynamic challenges scenes for RGBT tracking. 
In contrast, we employ dynamic routing methods to enable the attribute fusion branches to adapt their structure to the corresponding challenging scenarios at different levels of difficulty, and the fusion branches are utilized to disentangle the fusion progress to generate robust fusion features.

\begin{figure*}
\centering
\includegraphics[width=6.5in]{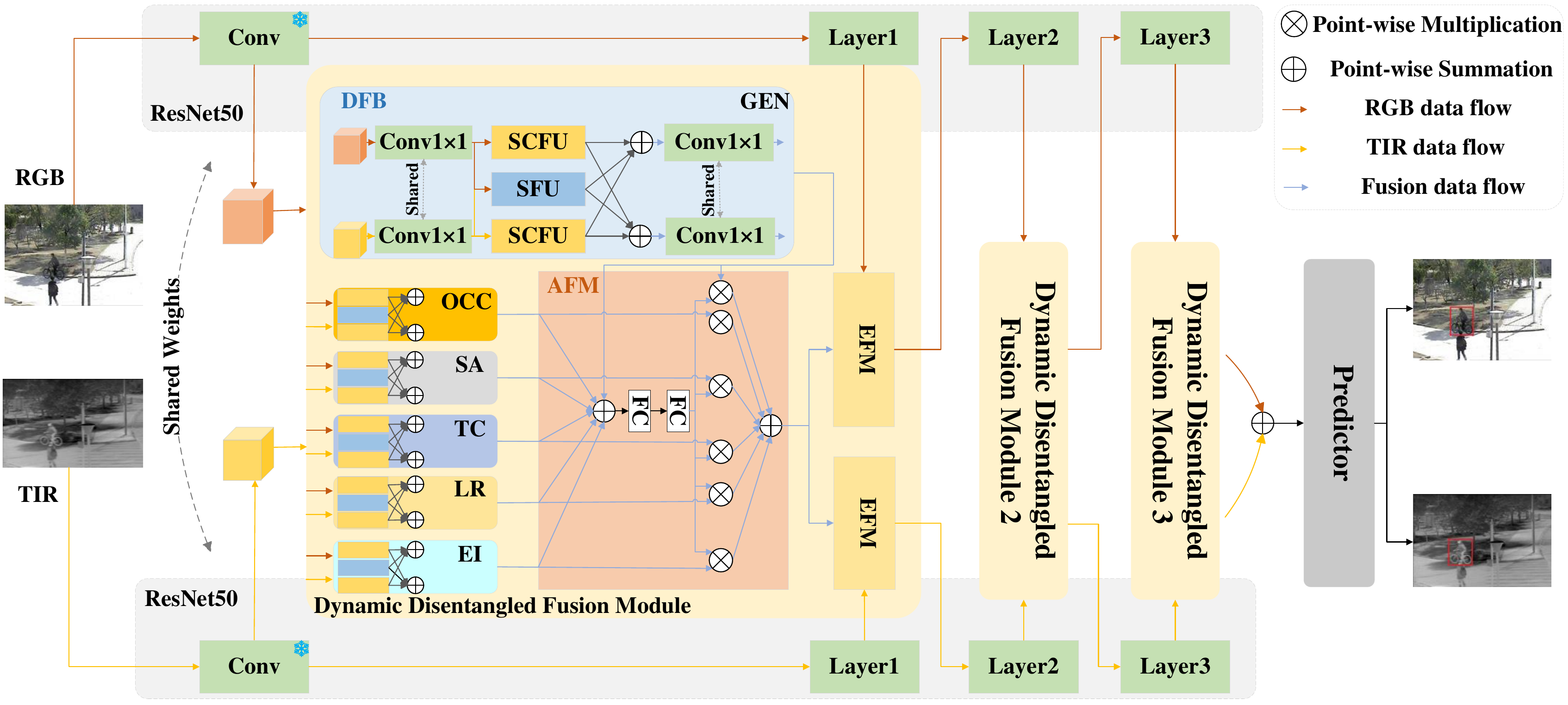}
\caption{
The proposed dynamic disentangled fusion network. 
The EFM denotes the lightweight enhancement fusion module. 
The acronyms IE, TC, OCC, LR, SA, and GEN stand for the dynamic fusion branches based on extreme illumination, thermal crossover, occlusion, low resolution, similar appearance, and general attributes respectively. 
The detailed structure of the Adaptive Aggregation Fusion Module (AFM) is shown in the network. 
}
\label{Overall framework}
\end{figure*}

\section{Methodology}
In this part, we first introduce the overall architecture of the Dynamic Disentangled Fusion Network (DDFNet). 
Second, the fundamental tracking network we use is shown in detail.
Third, we describe the design of each component of the dynamic disentangled fusion module, including six attribute-based dynamic fusion branches, an adaptive aggregation fusion module, and a lightweight enhancement fusion module.
Finally, the proposed three-stage training algorithm is described in detail.

\subsection{Overview}
In this section, we present the details of the overall architecture of DDFNet, in which the baseline tracker is ToMP~\cite{Mayer2022track}. The DDFNet is mainly composed of the Dynamic Disentangled Fusion (DDF) modules which include six dynamic fusion branches, an adaptive aggregation fusion module, and a lightweight enhancement fusion module. The detailed overall structure is shown in Figure~\ref{Overall framework}. 
Specifically, we use a two-stream feature extraction network constructed by ResNet-50~\cite{He2016resnet} to extract features from RGB and TIR images respectively. 
In each layer, we embed DDF modules in layer 1, layer 2, and layer 3 of the backbone network to gradually fuse the information from two modalities.

The main workflow is as follows.
First, the RGB and TIR images are sent to the backbone network for extracting modality-specific features separately, and the dynamic fusion branches perform feature fusion of all attribute branches simultaneously.
Then, all attribute-based fused features are sent to the adaptive aggregation module to obtain the aggregated features.
Next, the two modality-specific features and aggregated features are respectively sent to the lightweight enhancement fusion module to form robust feature representations, which are used as the input in the next convolution layer and DDF module. 
After the last DDF module, a Transformer-based predictor is used to extract global features for target classification and regression.

\subsection{Baseline Tracking Model}

The previous optimization-based trackers aim to solve an optimization problem where the target model generates the desired target state \(y_i\in\mathcal{Y}\) for the training samples \(\mathcal{S}_{\mathrm{train}} \in \{(x_{i},y_{i})\}_{i=1}^{n}.\) Here, \(x_i\in\mathcal{X}\) represents the deep feature map of frame \(i\) and \(n\) is the total number of training frames. The corresponding optimization problem is formulated as follows,
\begin{equation}
w=\arg\min_{\tilde{w}}\sum_{(x,y)\in\mathcal{S}_{\mathrm{train}}}f(h(\tilde{w};x),y)+\lambda g(\tilde{w}),
\end{equation}
where the objective function includes the residual function \(f\) which measures the error between the target model's output \(h((\tilde{w};x),y)\) and the ground truth label \(y\). The term \(g(\tilde{w})\) represents the regularization, scaled by a scalar \(\lambda\), while \(w\)  denotes the optimal weights of the target model. The training set \(\mathcal{S}_{\mathrm{train}} \) consists of the annotated first frame and previous tracked frames, with the tracker's predictions used as pseudo-labels. 
However, optimization-based methods rely on limited information from previously tracked frames to predict the target model.
%

To address the limitations of optimization-based target localization methods, ToMP~\cite{Mayer2022track} directly learns to predict the target model from data through end-to-end training, which contains a ResNet-based backbone network, a Transformer-based target model predictor, and a target model. In particular, ToMP first uses the backbone to extract testing features \( x_{\mathrm{test}} \in R^{H \times W \times C}\) and training features \( x_{i} \in R^{H \times W \times C} \), where \(x_{\mathrm{test}}\) is generated from the current frame. 
%
Then, the target state information is encoded into \( x_{i} \) to obtain the features \( v_{i} \) and \( x_{test} \) to obtain the features \( v_{test} \). 
Next, all training features \( v_{i} \) and the testing features \( v_{test} \) are concatenated along the first dimension and then simultaneously processed by a Transformer encoder \( T_{enc} \):

\begin{equation}\begin{aligned}
[z_{1},\ldots,z_{n},z_{\mathrm{test}}] = T_{enc}([v_{1},\ldots,v_{n},v_{\mathrm{test}}]).
\end{aligned}\end{equation}
The outputs of the Transformer encoder (\(z_{i}\) and \(z_{\mathrm{test}}\)) are used as inputs for the Transformer decoder \( T_{dec} \) to predict the target model weights and the operation can be represented as follow: 
\begin{equation}\begin{aligned}
w_t = T_{dnc}([z_{1},\ldots,z_{n},z_{\mathrm{test}}]),
\end{aligned}\end{equation}
where \( w_{t} \) indicates the weights of target model \(t\).

The output \( w_t \) of the Transformer decoder is fed into a linear layer to generate the weights for bounding box regression \( w_{t,bbreg} \) and target classification \( w_{t,cls} \). 
Based on \( w_{t,bbreg} \) and \( w_{t,cls} \), target classification and bounding box regression are performed by feeding enhanced testing features \(v_{\mathrm{test}}\) into the target model, resulting in predicted target classification \( \hat{y}_{test} \) and bounding box \( \hat{d}_{test} \). Please refer to~\cite{Mayer2022track} for more details.

\subsection{Dynamic Disentangled Fusion Module}
Existing attribute-based RGBT trackers~\cite{li2020rgbt, Zhang2021rgbt,liu2024rgbt} have achieved good performance. 
%
However, these trackers using a single fusion network cannot adapt well to various challenging scenarios and struggle to effectively integrate the features from two modalities.
Moreover, these attribute-based branches are designed with fixed structures, which are difficult to adapt to dynamic challenging environments over time.
As a result, they cannot effectively fuse features in challenge scenarios, thus limiting further improvements in tracking performance.
To address the above issues, we design the dynamic disentangled fusion module to extract the attribute-based fusion features.
In addition, existing work~\cite{Ma2015HierarchicalCF} has proved sufficient use of hierarchical features can enhance the tracking effectiveness, where the features extracted in the earlier layers have more detailed information to ensure the accurate localization of the target, and the features extracted in the latter have more semantic information to ensure the recognition of the target. 
Therefore, we adopt the hierarchical structure and insert dynamic disentangled fusion modules into the last three layers of the backbone network. The detailed design of the dynamic disentangled fusion module is shown in Figure~\ref{Overall framework}.

\begin{figure}

\centering
\includegraphics[width=3.5in]{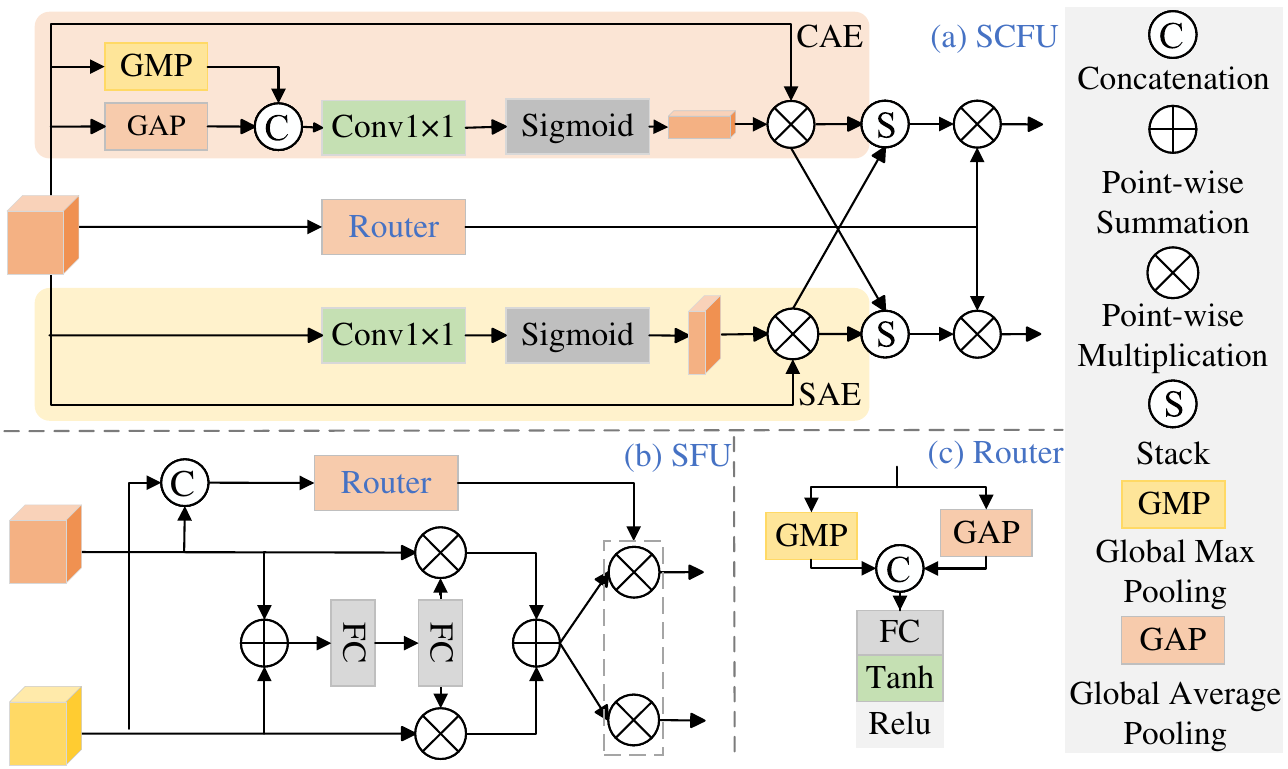}
\caption{ 
The dynamic fusion branch is comprised of the Spatial and Channel Fusion Unit (SCFU), Selective Fusion Unit (SFU), and a router. The structures of SCFU and SFU are shown in (a) and (b), and the structure of the router is shown in (c).  Herein, SCFU is composed of Spatial Attention Enhancement Module (SAE) and Channel Attention Enhancement Module (CAE), whose detailed designs are shown above.
}
\label{DFB}
\end{figure}

{\flushleft\textit{1) Dynamic Fusion Branch:} }
As shown in Figure~\ref{Overall framework}, we introduce five dynamic fusion branches (including thermal crossover (TC), extreme illumination (EI), similar appearance (SA), occlusion (OCC), and low resolution (LR)) to adequately fuse target features in various challenge scenarios. 
Then, to better address challenges beyond these five challenge attributes, we introduce a general (GEN) attribute dynamic fusion branch.
For each dynamic fusion branch, we can employ a small number of parameters to design the fusion branch since each branch is only responsible for a specific challenge attribute. Moreover, these branches can be trained separately using small-scale training data with the corresponding attributes without relying on large-scale training data. In particular, the structure of each dynamic fusion branch is composed of three router-guided fusion units, including two spatial and channel fusion units with the same structure and one selective fusion unit, and can be adaptively formed to adapt to dynamic challenge environments by combining these fusion units. 

Next, we first describe the spatial and channel fusion units in detail, which consist of a spatial attention enhancement module and a channel attention enhancement module. The unit can enhance the spatial and channel features of both modalities. 
Then, we show the selective fusion unit that adaptively selects the complementary information between the two modalities for dynamic fusion.
Finally, the structure of the router is described and it is explained how the guidance signals are generated.

{\flushleft\textbf{Spatial and Channel Fusion Unit:}} 
To adaptively select and fuse effective features for forming the robust representation of the target in challenging scenarios, we introduce the Spatial and Channel Fusion Unit (SCFU), which consists of a Spatial Attention Enhancement Module (SAE) and a Channel Attention Enhancement Module (CAE). The output features of the SAE and CAE are multiplied respectively by the weights generated from a router, and the results are applied to the modalities separately.


Specifically, to better capture target features and reduce the influence of target-irrelevant features, we introduce the spatial attention enhancement module to enhance the feature representations of the target and lower interference from target-irrelevant background regions, thus ensuring stable and accurate target tracking. The details of SAE are shown in Figure~\ref{DFB}(a). 
A convolution layer \(g(\cdot)\) with kernel size 1×1 and a sigmoid function \(\sigma(\cdot)\) are applied to generate spatial weights of the input features. 
The process of the proposed spatial attention enhancement module can be represented as follows:
\begin{equation}\begin{aligned}
\mathcal{F}_{SAE}^{l}(f_{m}^{l}) = \sigma(g(f_{m}^{l}))f_{m}^{l},
\end{aligned}\end{equation}
where \(\mathcal{F}_{SAE}^{l}\) denotes the SAE in the \(l\)-th layer, and \(f_{m}^{l}\) represents the features of modality \(m\) inputting the \(l\)-th layer of the backbone.

In addition, we design the channel attention enhancement module that reweights the feature channels so that the tracker pays more attention to interference-free feature channels, as shown in Figure~\ref{DFB}(a).
For example, in a video frame with the occlusion attribute, feature channels corresponding to the occluded region do not contain recognition information about the target, they may contain noise from the occluded region. 
Therefore, to effectively learn target-related features, we first obtain key information of the modal features \(f_{m}\) by the global average pooling layer \(GAP(\cdot)\) and the global max pooling layer \(GMP(\cdot)\). 
Then, we aggregate the key information with concatenation \(c(\cdot)\) and obtain the channel weights of the input features via a convolution layer and a sigmoid function. 
Finally, we weigh the feature channels using the element multiplication function.  
The process of the proposed channel attention enhancement module can be represented as follows:
\begin{equation}\begin{aligned}
\mathcal{F}_{CAE}^{l}(f_{m}^{l}) = \sigma(g(
c(GAP(f_{m}^{l}),GMP(f_{m}^{l})))
)f_{m}^{l},\\
\end{aligned}\end{equation}
where \(\mathcal{F}_{CAE}^{l}\) denotes the CAE in the \(l\)-th layer.

{\flushleft\textbf{Selective Fusion Unit:} }
To capture the complementary information of different modalities, we introduce a Selective Fusion Unit (SFU) based on SKNet, which adaptively selects and fuses channel-level features from two modal features. The structure of SFU is shown in Figure~\ref{DFB}(b).
Subsequently, the output features of SFU are multiplied by the weights generated from a router, and the results are applied to different modalities separately.
Specifically, we first aggregate the features of the two modalities via a global average pooling layer and a global max pooling layer. 
Then two fully connected layers for dimension expansion are used to obtain two feature vectors with the same dimensions. 
Next, we apply a softmax operation on the two feature vectors to obtain the channel weights of the features from both modalities. 
Finally, the channel weights are multiplied by the features from both modalities to obtain the final fused features. 

{\flushleft\textbf{Router:}} 
As shown in Figure~\ref{DFB}(c), each fusion unit includes a router that predicts the combination weights and determines whether the fusion unit should be combined with other units.
In particular, the router is implemented with two global pooling layers, a multi-layer perceptron, and two activation functions. 
Formally, the operation of the router in the $l$-th layer can be represented as follows:
\begin{equation}
\begin{aligned}
&\mathcal{R}_{SFU}^{l}(f_{rgb}^{l},f_{tir}^{l})&&= r(c(f_{rgb}^{l},f_{tir}^{l})),\\
&\mathcal{R}_{SCFU}^{l}(f_{m}^{l})&&=r(f_{m}^{l}),
\end{aligned}
\end{equation}
where \({R}_{SFU}^{l}\) and \({R}_{SCFU}^{l}\) indicatdes the routers of SFU and SCFU in the \(l\)-th layer. While the routers in SFU generate the bootstrap signal by both RGB \(f_{rgb}^{l}\) and TIR \(f_{tir}^{l}\) features simultaneously, the routers in the two SCFU generate the bootstrap signal by inputting modal features. The operation \(r(\cdot)\) on features \(f\) is as follows:
\begin{equation}
\begin{aligned}
r(f) =Relu(Tanh(MLP(c(GAP(f),GMP(f))))),
\end{aligned}
\end{equation}
where \(Relu\) and \(Tanh\) indicate Relu and Tanh activation functions, and \(MLP\) denotes the multilayer perceptron.

\begin{figure}
\centering
\includegraphics[width=0.48\textwidth]{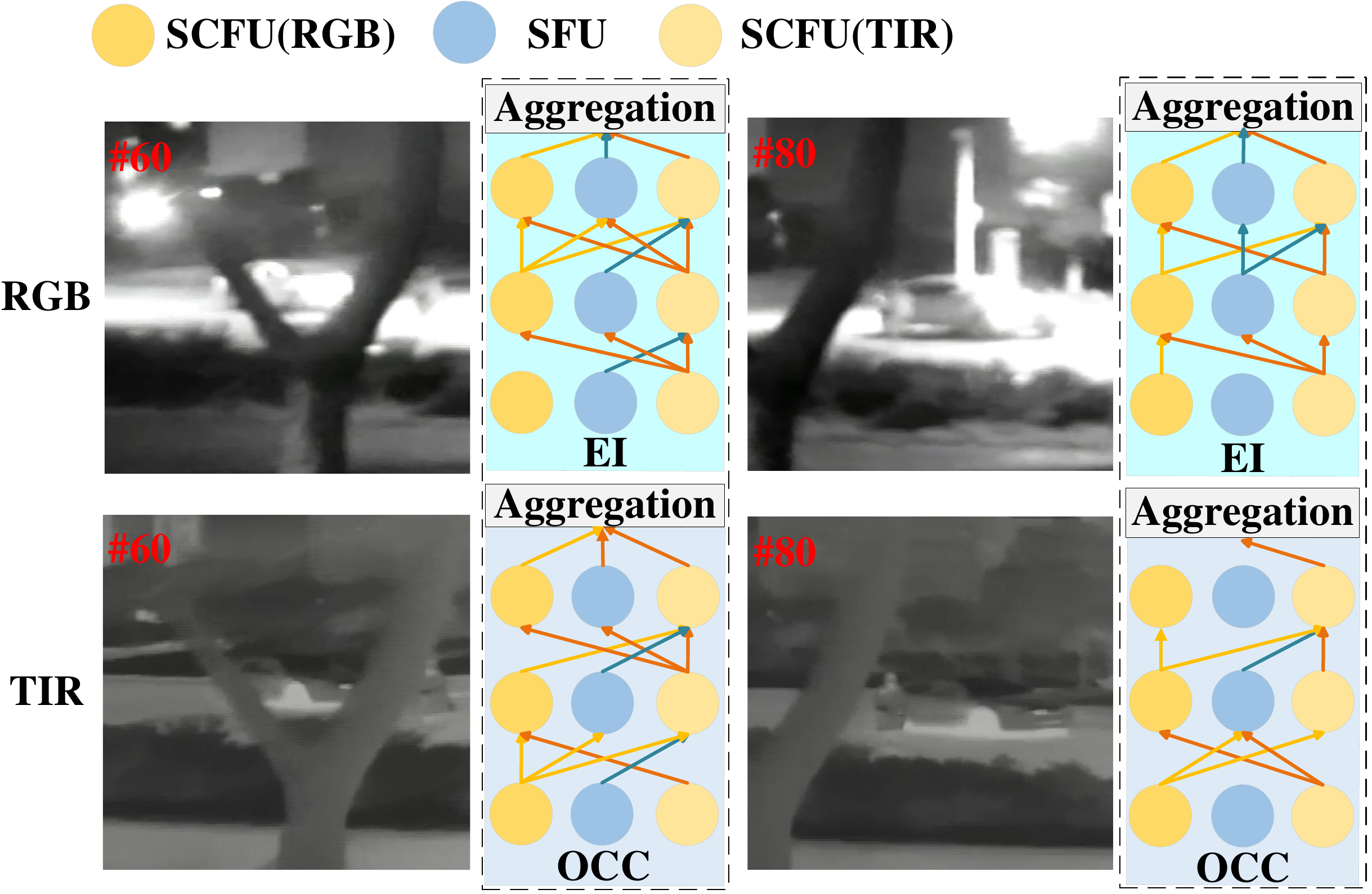}
\caption{ Visualization of the dynamic structure changes of the dynamic fusion branches in challenge scenarios.
}
\label{structure}
\end{figure}

To further validate the effectiveness of our dynamic fusion branch, we show the dynamic structure changes of the EI and OCC attribute branches in the shinycarcoming on the LasHeR dataset, as shown in Figure~\ref{structure}. As the tracking scene changes, the structure of the two attribute branches changes at the 80-th frame, where the OCC attribute branch changes significantly due to the disappearance of the occlusion challenge. 
This proves that our dynamic fusion branch can effectively adjust the network structure dynamically according to the challenge scenarios. 

\begin{figure*}[!t]
\centering
\includegraphics[width=1\textwidth]{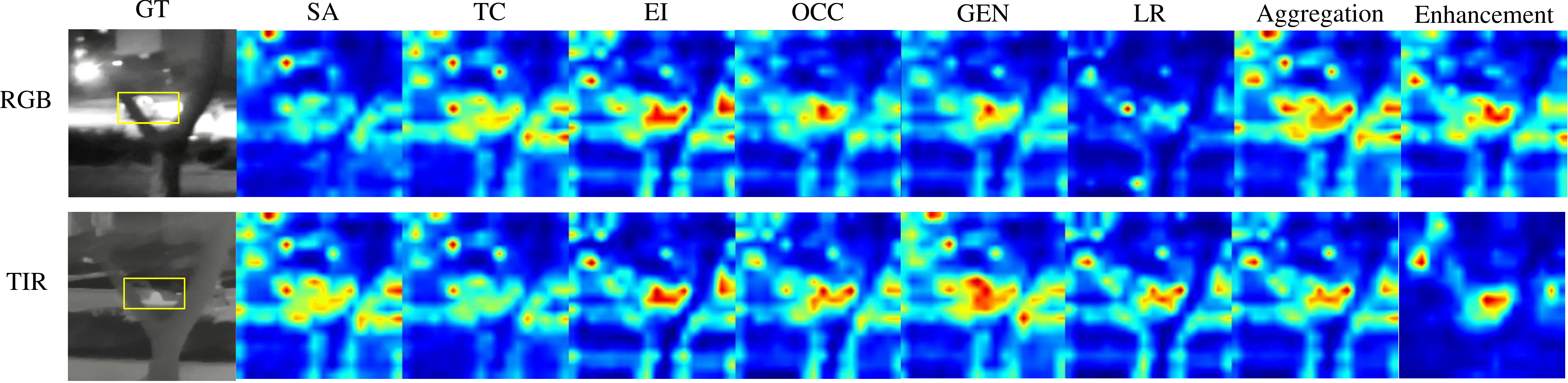}

\caption{ Feature map visualization of the attribute fusion features in dynamic fusion branches, the aggregated features in the adaptive aggregation fusion module, and the enhanced features in the lightweight enhancement fusion module. 
}
\label{Feature maps}
\end{figure*}

In addition, to display the quality of fusion of each attribute branch in a specific challenge scenario, we also visualize the fusion feature maps of all the attribute branches in Figure~\ref{Feature maps}. 
We can observe that all attribute branches are better able to achieve accurate target localization at the TIR branch compared to the RGB branch. Specifically, the EI attribute branch generates strong attention in both RGB and TIR branches. 
The OCC and GEN attribute branches also perform effective attention for the target.
In contrast, the SA, TC, and LR attribute branches do not generate such good attention compared to the other three branches. 
According to the above analysis, it can prove that our proposed dynamic fusion branches can effectively fuse features under the corresponding challenge attributes. 

{\flushleft\textit{2) Adaptive Aggregation Fusion Module:} }
Since it is uncertain which fusion branches should be activated during the tracking phase, we design an Adaptive Aggregation Fusion Module (AFM) that can dynamically combine features from all fusion branches.
Specifically, we first input all attribute-based fusion features into two fully connected layers, then pass through a softmax layer to obtain the feature channel weights of all the fused features.
Then, we perform a weighting operation on these six attribute-based fusion features with the weights obtained above to obtain more robust aggregated fusion features. The details are also shown in Figure~\ref{Overall framework}.

To demonstrate the important role of AFM in specific challenge scenarios, we show the feature maps of the aggregated features in Figure~\ref{Feature maps} (Aggregation), it can be seen that the adaptive aggregation fusion module generates robust aggregated features by aggregating all attribute-based fusion features. It proves that the AFM can adequately aggregate valid fusion features from all fusion branches.

\begin{figure}
\centering
\includegraphics[width=0.48\textwidth]{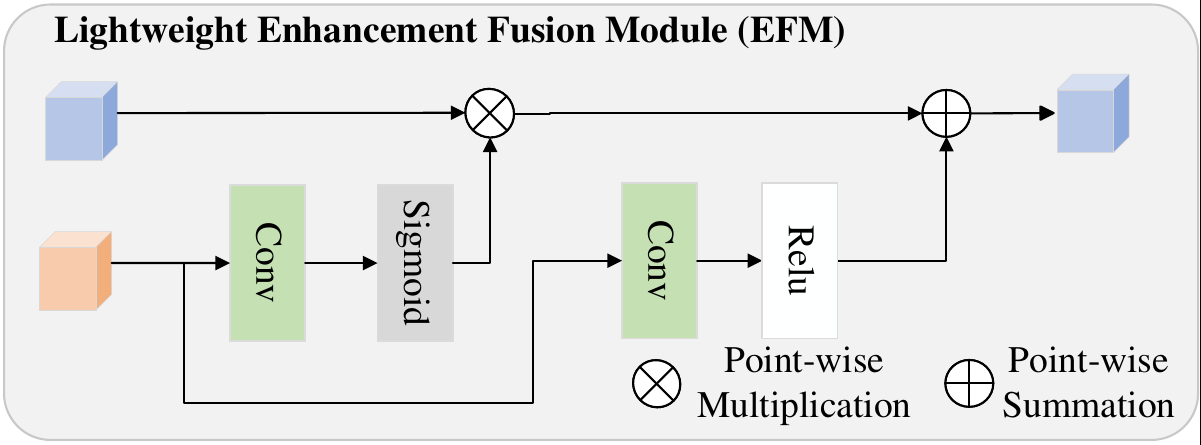}
        \caption{Overall structure of the Lightweight Enhancement Fusion Module (EFM). Each Conv module consists of a 1 × 1 convolution layer and a 3 × 3 convolution layer. The upper side input features are RGB or TIR features, while the lower side input features are aggregated features.}
\label{EFM}
\end{figure}

{\flushleft\textit{3) Lightweight Enhancement Fusion Module:} }
To fully achieve multimodal fusion and effectively utilize the information from both modalities, we propose an Lightweight Enhancement Fusion Module (EFM). Figure~\ref{EFM} shows the details of the lightweight enhancement fusion module.  
We feed the aggregated features and modality-specific features into the two parameter-independent lightweight enhancement fusion modules which enable comprehensive information interaction between modalities by employing aggregated features to guide the enhancement of the features from both modalities.
 
%
%
Specifically, the aggregated features first pass a convolutional layer and a sigmoid function to generate spatial weights to guide the modality-specific features for enhancement. 
Then, we send the aggregated features to a convolutional layer and a Relu function to suppress the spread of noise information in the aggregated features. 
Finally, the aggregated features and enhanced modal-specific features are fused, respectively.
The whole process can be formulated as:
\begin{equation}\begin{aligned}
\mathcal{F}_{EFM}^{l}(\hat{f}_{m}^{l},\hat{f}_{ag}^{l})=\hat{f}_{m}^{l}\sigma(g(\hat{f}_{ag}^{l}))+Relu(g(\hat{f}_{ag}^{l})),
\end{aligned}
\end{equation}
where \(\mathcal{F}_{EFM}^{l}\) indicates the EFM in the \(l\)-th layer, \(\hat{f}_{ag}^{l}\) denotes the aggregated features of \(l\)-th layer, and \(\hat{f}_{m}^{l}\) represents the features of modality \(m\) outputted by the \(l\)-th layer of the backbone.

As shown in Figure ~\ref{Feature maps} (Enhancement), the enhanced features further exclude the influence of background features compared to the aggregated features, making our network more focused on target related features. It indicates the effectiveness of EFM in further achieving full and effective fusion of both modalities.

\subsection{Three-stage Training Algorithm}
\begin{figure}[!b]
\centering
\includegraphics[width=0.48\textwidth]{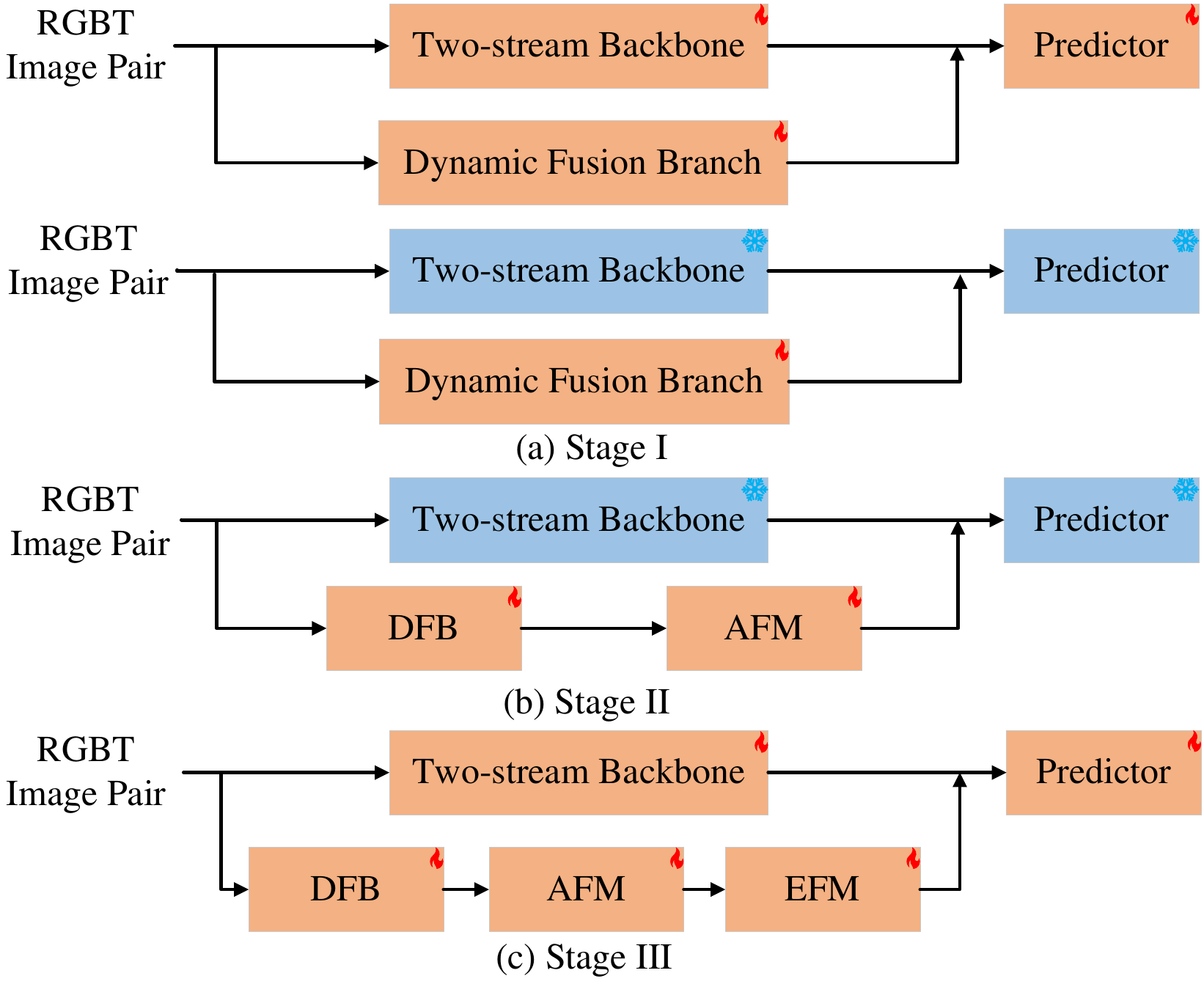}
\caption{ The visualization of our three-stage training algorithm.
}
\label{three-stage training}
\end{figure}

There are three key challenges to address during the training process. 
First, if the network is trained using all the training data at once, the loss from any attribute will be backpropagated through all dynamic fusion branches. 
Second, we do not know what attributes will appear in a frame and which dynamic fusion branches should be activated during the tracking phase. 
Finally, we want to enhance the fusion features of the dynamic fusion branch corresponding to the input data while suppressing the noise from other branches.
To address these issues, we propose a three-stage training algorithm to achieve both the effectiveness and efficiency of network training, as shown in Figure~\ref{three-stage training}. In addition, we generate data on the LasHeR dataset based on~\cite{liu2024rgbt} for training our dynamic disentangled fusion network.

{\flushleft\textit{1) Training of the all Dynamic Fusion Branches:} }
In the first stage, each dynamic fusion branch is trained individually. 
The adaptive aggregation fusion module and the lightweight enhancement fusion module are removed. The AdamW optimization strategy is used to optimize the network with the weights decay set to \(1\times10^{-4}\). 

First, we train the general attribute branch with the LasHeR dataset. Specifically, the model's two-stream ResNet 50 and Transformer-based predictor head are loaded with the pre-trained model parameters of ToMP50 for initialization. The learning rate for the general attribute fusion branch is set to \(1\times10^{-5}\) and the learning rate for the two-stream ResNet, as well as the Transformer prediction head, is set to \(5\times10^{-6}\). We set the training period to 30 epochs and save all module parameters.

Then the backbone and predictor of the model are loaded with the weight trained before, and the other dynamic fusion branches are trained one by one on the corresponding generated attribute-based training data. 
It is worth noting that when training a specific branch, the parameters of other branches are frozen.
In particular, we train only one fusion branch at a time and freeze the parameters of the other branches to ensure each fusion branch learns knowledge under a specific attribute.
The learning rate for the specific fusion branch is set to \(1\times10^{-5}\). We set the training period to 30 epochs and only save the parameters of dynamic fusion branches.

{\flushleft\textit{2) Training of the Adaptive Aggregation Fusion Module:}} In the second phase, we load all the parameters saved in the first stage and only train the adaptive aggregation fusion module using the LasHeR dataset. We randomly initialize the adaptive aggregation fusion module and set the learning rate to \(1\times10^{-5}\). The training period is 30 epochs. Other settings are the same as in the first stage. In this stage, we save all the parameters of the training model.

{\flushleft\textit{3) Training of the Lightweight Enhancement Fusion Module:} }
With the learned baseline tracker, six attribute branches, and adaptive aggregation fusion module obtained, we train the lightweight enhancement fusion module and fine-tune the other modules of our DDFNet on the LasHeR dataset. 
We set the learning rate of the parameters of the lightweight enhancement fusion module to \(1\times10^{-5}\) and other modules to \(1\times10^{-6}\). 
The training period is 60 epochs. The other settings are the same as in the second stage. In this stage, we save the parameters of the whole model.
\begin{table*}[ht]
	\centering
    \setlength{\tabcolsep}{2mm}{
	\caption{The PR, NPR, and SR scores (\%) of our DDFNet on GTOT, RGBT210, RGBT234, and LasHeR compared to various trackers. The best and second results are highlighted in $\textcolor{red} {red}$ and $\textcolor{blue}{blue}$ colors, respectively.}
    \label{overall_result}
    \renewcommand\arraystretch{1.4}{
	\resizebox{1\textwidth}{!}{
	\begin{tabular}{ccccccccccccc}
		\toprule
		\multirow{2}{*}{Methods} & \multirow{2}{*}{Publication} & \multirow{2}{*}{Backbone} & \multicolumn{2}{c}{GTOT} & \multicolumn{2}{c}{RGBT210} & \multicolumn{2}{c}{RGBT234} & \multicolumn{3}{c}{LasHeR} & FPS\\
        \cline{4-13}
		& & & PR$\uparrow$ & SR$\uparrow$ & PR$\uparrow$ & SR$\uparrow$ & PR$\uparrow$ & SR$\uparrow$ & PR$\uparrow$ & NPR$\uparrow$ & SR$\uparrow$ & $\uparrow$ \\
		\hline
        CAT~\cite{li2020rgbt} & ECCV 2020 & VGG$-$M & 88.9 & 71.7 & 79.2 & 53.3 & 80.4 & 56.1 & 45.0 & 39.5 & 31.4 &20 \\
        ADRNet~\cite{Zhang2021rgbt} & IJCV 2021 & VGG$-$M & 90.4 & 73.9 & $-$ & $-$ & 80.7 & 57.0 & $-$ & $-$ & $-$ &25 \\
        APFNet~\cite{Xiao2022rgbt} & AAAI 2022 & VGG$-$M & 90.5 & 73.7 & $-$ & $-$ & 82.7 & 57.9 & 50.0 & 43.9 & 36.2 &1.3 \\
        CAT$++$~\cite{liu2024rgbt} & TIP 2024 & VGG$-$M & \textcolor{blue}{91.5} & 73.3 & 82.2 & 56.1 & 84.0 & 59.2 & 50.9 & 44.4 & 35.6 & 14 \\ \hline
        
        MANet$++$~\cite{lu2021adapter} & TIP 2021 & VGG$-$M & 88.2 & 70.7 & $-$ & $-$ & 80.0 & 55.4 & 46.7 & 40.4 & 31.4 & 25.4 \\
        DMCNet~\cite{Lu2020DualityGatedMC} & TNNLS 2022 & VGG$-$M & 90.9 & 73.3 & 79.7 & 55.5 & 83.9 & 59.3 & 49.0 & 43.1 & 35.5 & 2.3 \\
        ProTrack~\cite{Yang2022PromptingFM} & ACM MM 2022 & ViT$-$B & $-$ & $-$ & $-$ & $-$ & 78.6 & 58.7 & 50.9 & $-$ & 42.1 & \textcolor{blue}{30} \\
        HMFT~\cite{Zhang2022VisibleThermalUT} & CVPR 2022 & ResNet$-$50 & 91.2 & 74.9 & 78.6 & 53.5 & 78.8 & 56.8 & $-$ & $-$ & $-$ \\
        MFG~\cite{Wang2021MFGNetDM} & TMM 2022 & ResNet$-$18 & 88.9 & 70.7 & 74.9 & 46.7 & 75.8 & 51.5 & $-$ & $-$ & $-$  & $-$\\
        DFNet~\cite{Peng2021DynamicFN} & TITS 2022 & VGG$-$M & 88.1 & 71.9 & $-$ & $-$ & 77.2 & 51.3 & $-$ & $-$ & $-$ & $-$\\
        DRGCNet~\cite{mei2023drgcnet} & IEEE SENS J 2023 & VGG$-$M & 90.5 & 73.5 & $-$ & $-$ & 82.5 & 58.1 & 48.3 & 42.3 & 33.8  & 4.9\\
        JTPMA~\cite{JTPMA2023} & INF FUSION 2023 & VGG$-$M & 90.7 & 75.1 & $-$ & $-$ & 80.3 & 56.2 & 53.8 & $-$ & 37.3  & 15.1\\
        CMD~\cite{Zhang2023EfficientRT} & CVPR 2023 & ResNet$-$50 & 89.2 & 73.4 & $-$ & $-$ & 82.4 & 58.4 & 59.0 & 54.6 & 46.4  & \textcolor{blue}{30}\\
    
        ViPT~\cite{Zhu2023VisualPM} & CVPR 2023 & ViT$-$B & $-$ & $-$ & $-$ & $-$ & 83.5 & 61.7 & 65.1 & $-$ & 52.5 & $-$\\
        TBSI~\cite{Hui2023BridgingSR} & CVPR 2023 & ViT$-$B & $-$ & $-$ & 85.3 & \textcolor{blue}{62.5} & 87.1 & 63.7 & 69.2 & 65.7 & 36.2 & \textcolor{red}{36.2}\\
        QAT~\cite{QAT2023} & ACM MM 2023 & ResNet$-$50 & \textcolor{blue}{91.5} & \textcolor{blue}{75.5} & \textcolor{blue}{86.8} & 61.9 & \textcolor{blue}{88.4} & \textcolor{blue}{64.4} & 64.2 & 59.6 & 50.1& 22  \\ \hline
        
        
        TATrack~\cite{TATrack} & AAAI 2024 & ViT$-$B & $-$ & $-$ & 85.3 & 61.8 & 87.2 & \textcolor{blue}{64.4} & \textcolor{blue}{70.2} & \textcolor{red}{66.7} & \textcolor{blue}{56.1}& 26.1  \\
        BAT~\cite{BAT2024} & AAAI 2024 & ViT$-$B & $-$ & $-$ & $-$ & $-$ & 86.8 & 64.1 & \textcolor{blue}{70.2} & $-$ & \textcolor{red}{56.3} & $-$\\
        OneTracker~\cite{OneTracker} & CVPR 2024 & ViT$-$B & $-$ & $-$ & $-$ & $-$ & 85.7 & {64.2} & 67.2 & $-$ & 53.8 & $-$ \\
        {Un-Track}~\cite{Un-Track} & CVPR 2024 & ViT$-$B & $-$ & $-$ & $-$ & $-$ & 84.2 & 62.5 & 66.7 & $-$ & 53.6 & $-$ \\
        SDSTrack~\cite{SDSTrack} & CVPR 2024 & ViT$-$B & $-$ & $-$ & $-$ & $-$ & 84.8 & 62.5 & 66.5 & $-$ & 53.1  & 20.9\\
  	\hline
        DDFNet &  $-$ & ResNet$-$50 & \textcolor{red}{91.6} & \textcolor{red}{76.3} & \textcolor{red}{87.7} & \textcolor{red}{63.4} & \textcolor{red}{90.0} & \textcolor{red}{66.9} & \textcolor{red}{71.1} & \textcolor{blue}{66.5} & 55.9 & 16 \\ \bottomrule
       \end{tabular}}}}

\end{table*}
\section{Experiments}

To evaluate the effectiveness of DDFNet, we compare our method with previous state-of-the-art methods on four RGBT tracking benchmark datasets, including GTOT~\cite{li2016rgbt}, RGBT210~\cite{li2017rgbt}, RGBT234~\cite{li2019rgbt}, and LasHeR~\cite{li2022rgbt}. In our experiments, we use the LasHeR dataset and the data generated based on the LasHeR dataset to train our DDFNet with the three-stage training algorithm.

\subsection{Evaluation Dataset and Metrics}

{\flushleft\textit{1) Evaluation Dataset:}} \textbf{GTOT} dataset is the first proposed RGBT tracking dataset, which contains 50 pairs of RGBT video sequences and 15K frames. To represent the performance of the RGBT tracker in a variety of challenges, the dataset is segmented into 7 subsets. \textbf{RGBT210} dataset is a large-scale RGBT tracking dataset. Compared to the GTOT dataset, this dataset contains 12 attributes in total. Accordingly, the amount of data increases to 210 pairs of RGBT video sequences with about 209K frames in total. \textbf{RGBT234} dataset is a superset of RGBT210, providing more accurate annotations while containing the same number of attributes. It includes 234 pairs of RGBT video sequences and a total of about 233K frames. \textbf{LasHeR} dataset is one of the biggest RGBT tracking datasets currently available, which comprises both a training set that includes 979 video pairs and a test set includes encompasses 245 video pairs. In total, this dataset encompasses 1224 video pairs and spans more than 1469K frames, adding 7 types of new attributes based on previous datasets, making it more challenging.

{\flushleft\textit{2) Evaluation Metrics:}} As for GTOT, RGBT210, and RGBT 234, the results are evaluated with maximum success rate (SR) and maximum precision rate (PR) via the pass evaluation rule as evaluation metrics. SR indicates the percentage of successfully tracked frames where the overlap between the tracking results and the ground truth is greater than a designated threshold. We derive SR by assessing the area under the curve and select the maximum value from the two modes as the final outcome. PR represents the maximum frame ratio, whose center location error between the prediction and ground truth is smaller than the threshold. The threshold is set to 5 pixels in the GTOT and 20 pixels in the other three datasets. Additionally, recognizing that the PR metric is highly sensitive to target size variation, The LasHeR dataset presents a normalized precision rate (NPR) for evaluating tracking performance. This NPR is calculated by adjusting the precision rate (PR) based on the size of the ground truth.

\subsection{Quantitative Comparison}
We test our DDFNet on four popular RGBT tracking benchmarks and compare performance with some state-of-the-art trackers, such as CAT~\cite{li2020rgbt}, ADRNet~\cite{Zhang2021rgbt}, APFNet \cite{Xiao2022rgbt}, CAT++~\cite{liu2024rgbt}, MANet++~\cite{lu2021adapter},  DMCNet~\cite{Lu2020DualityGatedMC}, ProTrack~\cite{Yang2022PromptingFM}, HMFT~\cite{Zhang2022VisibleThermalUT}, MFG~\cite{Wang2021MFGNetDM}, DFNet~\cite{Peng2021DynamicFN}, DRGCNet~\cite{mei2023drgcnet}, CMD~\cite{Zhang2023EfficientRT}, ViPT~\cite{Zhu2023VisualPM}, TBSI~\cite{Hui2023BridgingSR}, QAT~\cite{QAT2023},  TATrack~\cite{TATrack}, BAT~\cite{BAT2024}, OneTracker~\cite{OneTracker}, Un-Track~\cite{Un-Track}, SDSTrack~\cite{SDSTrack}, to validate the effectiveness of proposed method. 

{\flushleft\textit{1) Evaluation on GTOT Dataset:} }Comparison results on GTOT dataset are shown in Table~\ref{overall_result}. We can observe that our DDFNet achieves the best results with 91.6\%/76.3\% in PR/SR. In particular, our tracker achieves 1.2\%/2.4\%, 1.1\%/2.6\%, and 2.4\%/2.9\% improvements against ADRNet, APFNet, and CMD in PR/SR, respectively. We further compare our method with state-of-the-art trackers CAT++ and QAT, and our DDFNet surpasses CAT++ and QAT with 3.0\% and 0.8\% in SR, indicating superior performance in target scale regression.

{\flushleft\textit{2) Evaluation on RGBT210 Dataset:}} As shown in Table~\ref{overall_result}, our algorithm achieves the best tracking performance on RGBT210 dataset compared to all state-of-the-art trackers. Compared to the attribute-based trackers CAT and CAT++, DDFNet achieves substantial improvements in PR/SR with gains of 8.5\%/2.8\% and 5.5\%/7.3\%. Moreover, compared to QAT, the best performing algorithm on RGBT210 dataset, our method exhibits 1.9\%/1.5\% performance gains in PR/SR.

{\flushleft\textit{3) Evaluation on RGBT234 Dataset:}} We conduct experiments
on RGBT234 dataset and comparison with 20 RGBT trackers. As can be seen from Table~\ref{overall_result}, our algorithm achieves the best tracking performance with PR/SR scores of 90.0\%/66.9\% on RGBT234 dataset compared to all state-of-the-art trackers. It is worth noting that our method obtains significant performance gains compared to attribute-based trackers ADRNet, APFNet, and CAT++ with 9.3\%/9.9\%, 7.3\%/9.0\%, and 6.0\%/7.7\% in PR/SR. 
Compared with the state-of-the-art RGBT trackers TATrack and BAT, our DDFNet obtains significant improvements of 2.8\%/2.5\% and 3.2\%/ 2.8\% in SR/PR, respectively.
Besides, compared with QAT which is the top advanced tracker on RGBT234 dataset, our tracker outperforms it with 1.6\%/2.5\% in PR/SR respectively. These results fully demonstrate the effectiveness of our method.

{\flushleft\textit{4) Evaluation on LasHeR Dataset:}} The evaluation results on LasHeR testing set are shown in Table~\ref{overall_result}. Compared with 16 RGBT trackers, we can find that our tracker achieves excellent performance. In particular, our tracker achieves 3.9\%/2.1\%, 4.4\%/2.3\%, and 4.6\%/2.8\% improvements against OneTracker, Un-Track, and SDSTrack in PR/SR, respectively.
Compared to the state-of-the-art RGBT tracker TATrack, which utilizes ViT-B as the backbone network, our DDFNet still achieves a 0.9\% improvement in PR metrics and comparable performance in NPR/SR, even though we use ResNet-50 as the backbone network.
TATrack enhances the robustness of the tracker by leveraging temporal information. In contrast, our DDFNet focuses on using dynamic fusion branches to improve the fusion of effective features under different challenge attributes. 
It is worth noting that on the RGBT210 and RGBT234 datasets, our DDFNet significantly outperforms TATrack, indicating that our DDFNet has stronger generalization ability.
Compared with QAT which uses the same backbone network as our DDFNet, our DDFNet obtains significant improvements of 6.9\%/6.9\%/5.8\% in PR/NPR/SR, respectively.
QAT enhances high-quality modality features by predicting the quality of each modality to obtain feature fusion weights for multimodal integration.
However, relying solely on the predicted quality weights to enhance effective modality features limits the ability to fully utilize modality information.
In contrast, our DDFNet effectively integrates useful modality features during the feature extraction process by utilizing dynamic disentangled fusion modules.
Finally, compared with our conference paper APFNet, the improved method obtains significant improvements of 20.2\%/22.6\%/19.7\% in PR/NPR/SR.
The above results fully demonstrate the effectiveness of our DDFNet.

\begin{figure}
\centering
\includegraphics[width=0.4\textwidth]{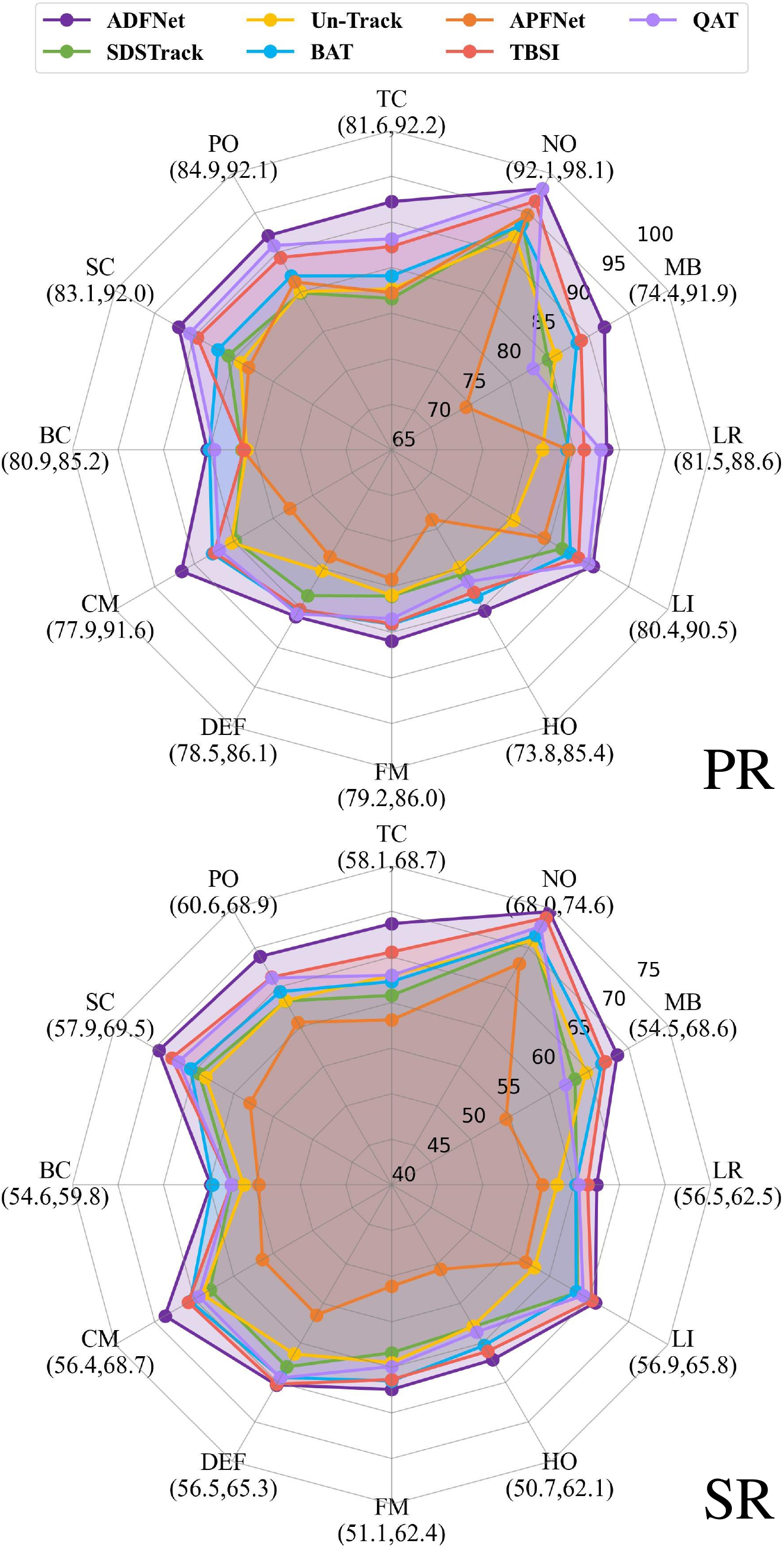}
\caption{ Precision Rate (PR) and Success Rate (SR) of challenge attributes on the RGBT234 dataset.}
\label{radar}
\end{figure}

{\flushleft\textit{5) Attribute-based Performance Evaluation:}} 
To further validate the advantages of our approach in different challenge attributes, we compare our DDFNet with other state-of-the-art RGBT trackers including  Un-Track, APFNet, SDSTrack, BAT, QAT, TBSI on the subsets of different challenging attributes on RGBT234 dataset. The challenge attributes include thermal crossover (TC), partial occlusion (PO), heavy occlusion (HO), low illumination (LI), low resolution (LR), distortion (DEF), scale variations (SV), motion blur (MB), no occlusion (NO), camera movement (CM), background clutter (BC), and fast motion (FM). The evaluation results are shown in Figure~\ref{radar}.

As can be seen from the results, our DDFNet achieves the best results on the 12 challenge attributes, which prove the excellent performance of our DDFNet in challenging scenarios. In particular, our DDFNet significantly outperforms the attribute-based tracker APFNet in all challenging attributes, especially in CM, HO, and MB where PR/SR metrics improve by 13.7\%/12.3\%, 11.6\%/11.4\%, and 17.5\%/14.1\%, respectively. Compared with the state-of-the-art method TBSI, our DDFNet also achieves superior performance in all attributes. Moreover, our DDFNet significantly outperforms it in the challenge attributes of BC, CM, and TC with improvements of 4.0\%/2.2\%, 4.2\%/3.0\%, and 4.9\%/3.2\% respectively. It demonstrates that our DDFNet can better fuse features in challenge attributes.

\begin{figure*}
\centering
\includegraphics[width=1\textwidth]{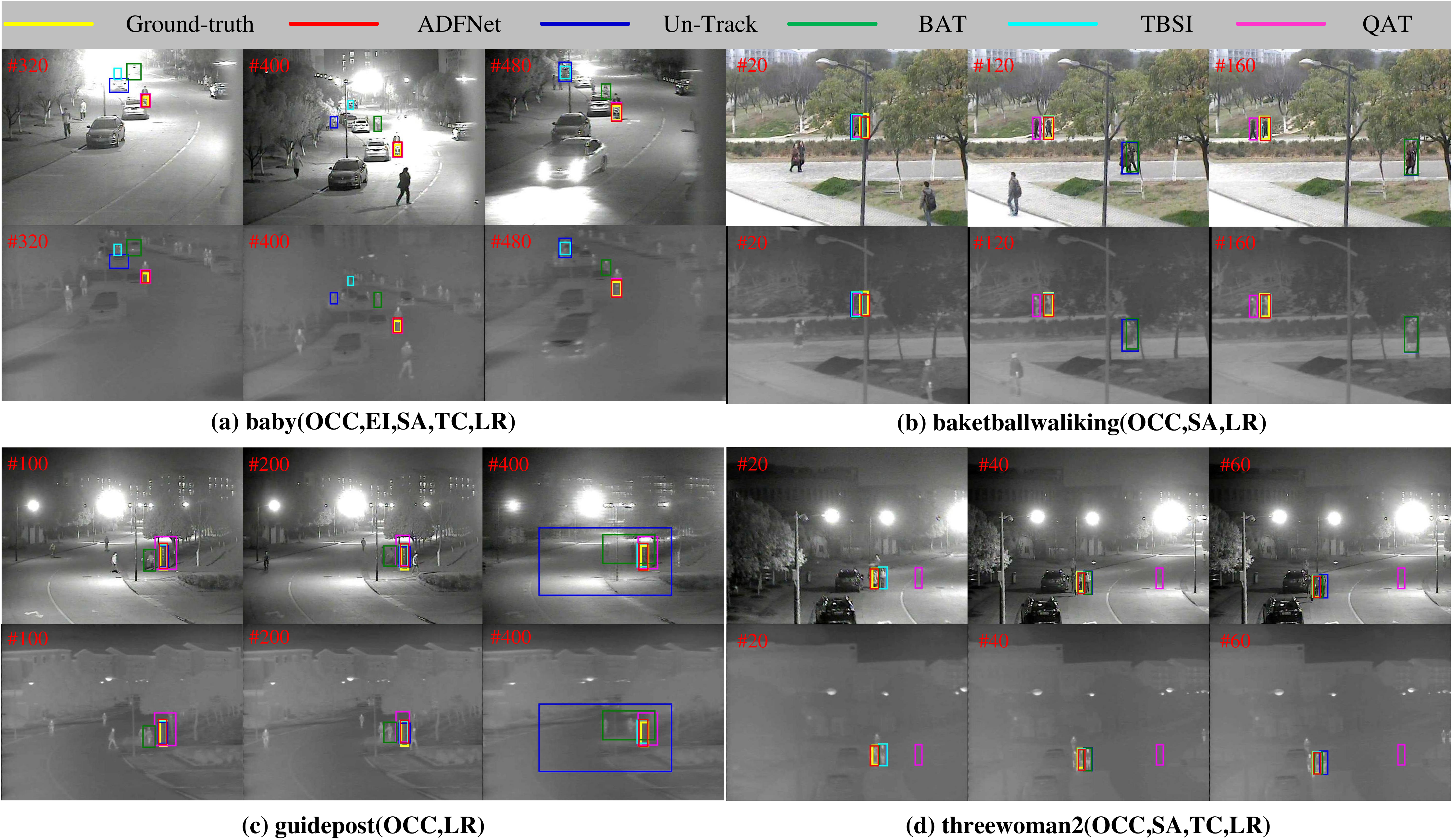}
\caption{ Qualitative comparison of DDFNet against four state-of-the-art trackers on four video sequences from RGBT234 dataset.}
\label{visualize}
\end{figure*}

{\flushleft\textit{6) Visual Comparison:}} 
As shown in Figure~\ref{visualize}, we visualize the tracking results of our DDFNet with state-of-the-art RGBT trackers Un-Track, BAT, TBSI, and QAT in video frames to intuitively validate the effectiveness of DDFNet. 
The results indicate that our DDFNet outperforms other trackers significantly when faced with challenge attributes.
For instance, when confronted with a small target in Figure~\ref{visualize} (baby), only our method correctly tracks the target in frames 400 and 480, and
both BAT and TBSI incorrectly track a nearby target.
While in the 20-th frame from basketballwalking sequence in Figure~\ref{visualize} which presents an occlusion scenario, our method tracks the target stably, whereas the other methods exhibit tracking errors.
The above analysis and visualization results fully demonstrate the effectiveness of our proposed method in being able to effectively perceive different attribute challenges as well as enhance the target features of the corresponding attribute branches.

\subsection{Ablation Study}
\begin{table}[ht]
	\centering
    \setlength{\tabcolsep}{2mm}
    \caption{Analysis of dynamic fusion branches across challenge attribute data on RGBT234 dataset. The best results are indicated in \textcolor{red}{red}.}
    \label{Dynamic Fusion Branches}
    \renewcommand\arraystretch{1.5}
    {
	\resizebox{0.48\textwidth}{!}{
	\begin{tabular}{cccccc}
		\toprule
		\multirow{2}{*}{Method} & \multirow{2}{*}{OCC Data} & \multirow{2}{*}{LR Data} & \multirow{2}{*}{SA Data} & \multirow{2}{*}{TC Data} & \multirow{2}{*}{EI Data} \\
         & & & & & \\
		\hline
        baseline & 83.9/61.9 & 84.9/56.0 & 54.5/40.6 & 83.6/56.9 & 71.5/51.9 \\
        GEN & 86.2/64.1 & 86.5/57.3 & 57.3/42.6 & 84.0/58.1 & 73.3/53.3 \\
        OCC & \textcolor{red}{94.0/70.6} & 74.0/49.1 & 38.7/28.8 & 83.8/58.8 & 68.7/49.9 \\
        LR & 83.0/61.8 & \textcolor{red}{86.6/58.6} & 56.5/42.5 & 82.7/59.3 & 67.6/48.7  \\
        SA & 75.8/53.3 & 82.6/53.8 & \textcolor{red}{94.1/72.2}& 79.7/54.4 & 66.9/48.1    \\
        TC & 85.3/63.2 & 85.8/57.1 & 57.5/42.5 & \textcolor{red}{88.2/65.7} & 67.9/49.2\\
        EI & 83.9/61.5 & 84.3/54.7 & 57.3/42.4 & 76.0/48.2 & \textcolor{red}{85.3/62.7} \\
        \bottomrule
       \end{tabular}}}
\end{table}
{\flushleft\textit{1) Analysis of Dynamic Fusion Branches:} }
To further quantitatively analyze the effectiveness of each dynamic fusion branch which is specific and capable of solving the corresponding challenge, we show a comparison of attribute-based data generated on RGBT234 dataset as shown in Table~\ref{Dynamic Fusion Branches}. We can find that the tracker with a general dynamic fusion branch achieves significant performance improvements over baseline ToMP50+RGBT on all attributes, outperforming 2.3\%/2.2\%, 1.6\%/1.3\%, 2.8\%/2.0\%, 0.4\%/1.2\%, and 1.8\%/1.4\% under OCC, LR, SA, TC, and EI challenge attributes, respectively. It indicates that the introduction of the general branch contributes well to the performance of the network on all challenge scenarios, fulfilling the expectations of this branch to cope with a wider range of challenge attributes. 
Besides, compared to the baseline tracker, all trackers equipped with an dynamic fusion branch achieve the best performance on the corresponding challenge attribute, outperforming 10.1\%/8.7\%, 1.7\%/2.6\%, 39.6\%/31.6\%, 4.6\%/8.8\%, and 13.8 \%/10.8\% under the challenge attributes of OCC, LR, SA, TC, and EI, respectively. These results suggest that designing branches for attributes can fully learn the information of the corresponding challenge attribute and effectively address corresponding challenges.

\begin{table}[ht]
	\centering
    \setlength{\tabcolsep}{2mm}{
	\caption{Analysis of fusion units. \ding{51} means that the unit is included in the dynamic fusion branch. The best results are indicated in \textcolor{red}{red}.}
    \label{Unit}
    \renewcommand\arraystretch{1.5}{
	\resizebox{0.48\textwidth}{!}{
	\begin{tabular}{c|ccc|ccccc}
		\toprule
		 \multirow{2}{*}{Methods} &\multirow{2}{*}{SCFU(RGB)} & \multirow{2}{*}{~~~SFU~~~} & \multirow{2}{*}{SCFU(TIR)} & \multicolumn{2}{c}{RGBT234} & \multicolumn{3}{c}{LasHeR}\\
        \cline{5-9}
         & & & & PR & SR & PR & NPR & SR \\
         \hline
         baseline & & & & 87.2 & 65.3 & 65.1 & 60.7 & 51.2 \\
         &\ding{51} & & & 88.1 & 65.9 & 67.8 &63.4 &  53.4\\
        
        & \ding{51}  &  \ding{51} &  & 89.0 & 66.3 & 70.0 & 65.3 & 54.8\\
        DDFNet & \ding{51}  & \ding{51} & \ding{51} & \textcolor{red}{90.0} & \textcolor{red}{66.9} & \textcolor{red}{71.1} & \textcolor{red}{66.5} & \textcolor{red}{55.9}\\
        \bottomrule
       \end{tabular}}}}
\end{table}

{\flushleft\textit{2) Analysis of Fusion Units:}}
We evaluate each fusion unit on both RGBT234 and LasHeR datasets, verify the effectiveness of the units, and summarize the results in Table~\ref{Unit}. As seen in row 2, directly using the SCFU unit individually also improves the performance of the network compared to the baseline ToMP50+RGBT, proving the effectiveness of the SCFU fusion unit and attribute-based fusion method. Then, as shown in row 3, the combination of two fusion units can further improve the tracking capability by 0.9\%/0.4\% on RGBT234 dataset and 1.2\%/0.9\%/0.6\% on LasHeR dataset, proving the effectiveness of the SFU fusion unit and the dynamic fusion structure. Especially in the last row, where all fusion units are added, the highest performance is achieved.

\begin{table}[ht]
	\centering
    \setlength{\tabcolsep}{2mm}{
	\caption{Analysis of AFM and EFM. \ding{51} means that the module is included in the dynamic disentangled fusion module. The best results are indicated in \textcolor{red}{red}.}
    \label{AFM EFM}
    \renewcommand\arraystretch{1.5}{
	\resizebox{0.48\textwidth}{!}{
	\begin{tabular}{c|ccc|ccccc}
		\toprule
		  \multirow{2}{*}{Methods} & \multirow{2}{*}{DFB} & \multirow{2}{*}{AFM} & \multirow{2}{*}{EFM}  & \multicolumn{2}{c}{RGBT234} & \multicolumn{3}{c}{LasHeR}\\
        \cline{5-9}
         & & & & PR & SR & PR & NPR & SR\\
         \hline
        baseline & & &  & 87.2 & 65.3 & 65.1 & 60.7 & 51.2\\
        &\ding{51}&  &   & 87.8  & 65.5  &  67.5 & 62.9  & 53.3 \\
        &\ding{51}& \ding{51} &   & 89.3  &  66.6 & 70.2 & 65.6 & 55.2 \\
        DDFNet & \ding{51}& \ding{51} & \ding{51} & \textcolor{red}{90.0} & \textcolor{red}{66.9} & \textcolor{red}{71.1} & \textcolor{red}{66.5} & \textcolor{red}{55.9}  \\
        
        \bottomrule
       \end{tabular}}}}
\end{table}

{\flushleft\textit{3) Analysis of AFM and EFM:}}
To validate the effectiveness of AFM and EFM, we test the performance of our DDFNet by adding AFM and EFM one by one, with comparative results shown in Table~\ref{AFM EFM} on RGBT234 and LasHeR datasets.
First, AFM and EFM are removed simultaneously, and the fusion features from all dynamic fusion branches are directly summed to modal features. However, the performance improvement is slight compared to the baseline ToMP50+RGBT, which indicates that aggregating the features from the dynamic fusion branches by summation is not effective. Second, AFM is added and the fusion features from all dynamic fusion branches are aggregated via AFM. The tracking performance is significantly improved. Finally, EFM is also added and the performance is further improved by fully fusing the modal features with the aggregated features through EFM. The results show that the increase of both AFM and EFM improves the tracking performance, which proves the effectiveness of AFM and EFM.

\begin{table}[ht]
	\centering
    \setlength{\tabcolsep}{2mm}{
	\caption{Analysis of hierarchical design. \ding{51} means adding a dynamic disentangled fusion module to this layer of backbone in ToMP50. The best results are indicated in \textcolor{red}{red}.}
    \label{Hierarchical Design}
    \renewcommand\arraystretch{1.4}{
	\resizebox{0.48\textwidth}{!}{
	\begin{tabular}{c|ccc|ccccc}
		\toprule
		  \multirow{2}{*}{Methods} & \multirow{2}{*}{layer1} & \multirow{2}{*}{layer2} & \multirow{2}{*}{layer3} & \multicolumn{2}{c}{RGBT234} & \multicolumn{3}{c}{LasHeR}\\
        \cline{5-9}
        & & & & PR & SR & PR & NPR & SR \\
         \hline
        baseline & & &  & 87.2 & 65.3 & 65.1 & 60.7 & 51.2\\
        & \ding{51} & & & 89.1 & 66.5 & 68.1 & 63.9 & 53.9 \\
        & \ding{51}  &  \ding{51} &  & 89.2 & 66.5 & 68.6 & 64.1 & 54.0 \\
        DDFNet & \ding{51}  & \ding{51} & \ding{51} &  \textcolor{red}{90.0} & \textcolor{red}{66.9} & \textcolor{red}{71.1} & \textcolor{red}{66.5} & \textcolor{red}{55.9}\\
        \bottomrule
       \end{tabular}}}}
\end{table}

{\flushleft\textit{4) Analysis of Hierarchical Design:}}
Incorporating dynamic disentangled fusion modules at every level of the backbone network would considerably raise computational complexity, potentially impacting tracking speed. Nonetheless, we contend that a hierarchical design is essential, as various challenge attributes are represented differently across the layers. To demonstrate our proposed hierarchical design is effective, we gradually insert dynamic disentangled fusion modules in one, two, and all layers of the backbone network, and the comparison results are shown in Table~\ref{Hierarchical Design}. From the results, we can observe that compared to the baseline ToMP50+RGBT, as the number of inserted layers increases the performance of the tracker shows a higher improvement on both RGBT234 and LasHeR datasets, which proves the effectiveness of hierarchical design for promoting fusion ability.

\section{Conclusion}
In this paper, we propose the novel Dynamic Disentangled Fusion Network (DDFNet) to perform effective fusion of different modalities via challenge attributes. First, we design the dynamic fusion branches, which can dynamically form a fusion structure by selecting fusion units based on the current challenge scenario. Then, the adaptive aggregation fusion module is introduced to aggregate all attribute-based fusion features. Finally, the lightweight enhancement fusion module is introduced to enhance aggregated features and modality-specific features. Extensive experiments on four benchmark datasets demonstrate the effectiveness of our method against state-of-the-art trackers. In the future, we plan to explore more efficient dynamic fusion structures to address more challenging attributes and to improve the efficiency of the fusion structure selection mechanism. Furthermore, since some challenges such as fast motion and scale variation require temporal information mining, we plan to investigate embedding temporal information into attribute-based branches. In addition, we will explore the paradigm of attribute-based approaches with advanced Transformer~\cite{ye2022joint} and Mamba~\cite{Wang2024} models as the backbone networks.

\bigskip

\bibliography{Dynamic_Disentangled_Fusion_Network_for_RGBT_Tracking}

\end{document}